\documentclass{article}
\usepackage{longtable}
\usepackage{array} 
\usepackage{float}
\usepackage{amsmath}
\usepackage{PRIMEarxiv}
\usepackage{makecell}
\usepackage[utf8]{inputenc} 
\usepackage[T1]{fontenc}    
\usepackage{hyperref}       
\usepackage{url}            
\usepackage{booktabs}       
\usepackage{amsfonts}       
\usepackage{nicefrac}       
\usepackage{microtype}      
\usepackage{lipsum}
\usepackage{fancyhdr}       
\usepackage{graphicx}       
\graphicspath{{media/}} 
\usepackage{xcolor}
\usepackage{authblk}
\usepackage{hyperref}
\usepackage{multirow} 
\usepackage{longtable} 
\usepackage{enumitem}
\usepackage{multirow}
\usepackage{array}
\usepackage{verbatim}
\begin{document}

\title{RESPClinBench: Benchmarking Multimodal Clinical Decision-Making and Longitudinal Disease Management in Respiratory Specialty Care}
\author[1]{Mouxiao Bian}
\author[2]{Zhi Chen}
\author[1]{Ruiyao Chen}
\author[1]{Lu Lu}
\author[3,4]{Hengrui Liang}
\author[3]{Chaoyi Huang }
\author[4]{Yiluo Lin}
\author[1]{Jingru Ding}
\author[1]{Yun Zhong}
\author[3,4,*]{Yueming Su}
\author[1,*]{Jie Xu}
\affil[1]{\textit{
   Shanghai Artificial Intelligence Laboratory \\
    Shanhai, China
}}
\affil[2]{\textit{
    Macau University of Science and Technology\\
    Macau, China
}}
\affil[3]{\textit{
    First Affiliated Hospital of Guangzhou Medical University \\
    Guangzhou, China
}}
\affil[4]{\textit{
   Guangzhou Institute of Respiratory Health \\
    Guangzhou, China
}}

\maketitle
 
\footnotetext[1]{*Correspondence: 
Jie Xu (xujie@pjlab.org.cn),
Yueming  Su(su\_yueming@gzlab.ac.cn)
}

\begin{abstract}
\textbf{Background: }Respiratory specialty care requires multimodal interpretation, longitudinal risk assessment, guideline-concordant intervention, and whole-course management, which are poorly represented by examination-oriented medical benchmarks.

\textbf{Objective: }To develop RESPClinBench, a real-world scenario-based benchmark for respiratory clinical decision-making, and evaluate seven contemporary large language models across AECOPD-PIM and PNBIM.

\textbf{Methods:} RESPClinBench cases were adapted from de-identified respiratory clinical data. Three attending-level respiratory physicians revised cases, reference answers, and atomic clinical-action points, while one senior respiratory specialist performed cross-review and final adjudication. AECOPD-PIM comprised 427 open-ended COPD cases, and PNBIM comprised 196 multimodal pulmonary nodule cases combining chest CT with structured clinical information. Seven models generated 4,361 responses through standardized API inference with temperature 0 and a maximum output length of 8192 tokens. An automated framework calculated the final score as the arithmetic mean of atomic-action recall and rubric-based LLM-as-a-Judge assessment.

\textbf{Results}: Across 623 cases, the mean final score was 68.58. Qwen3.6-27B ranked first overall at 71.22, Qwen3.5-397B-A17B led PNBIM at 72.48, and Qwen3.6-27B led AECOPD-PIM at 71.11. Imaging hallucination and serious medical risk occurred in 31.85\% and 8.16\% of PNBIM responses; medication-safety risk and serious medical risk occurred in 26.93\% and 1.44\% of AECOPD-PIM responses.

\textbf{Conclusions}: RESPClinBench identifies task-specific limitations in multimodal pulmonary nodule assessment and longitudinal COPD management. Combining explicit clinical-action coverage, holistic evaluation, and independent safety flags provides a clinically grounded basis for model selection and prospective validation.

\end{abstract}


\section{Introduction}
Large language models (LLMs) have rapidly progressed from general-purpose text generators to systems capable of answering medical questions, synthesizing clinical records, generating differential diagnoses, and participating in simulated consultations. Landmark evaluations such as MultiMedQA showed that language models can encode substantial clinical knowledge and approach expert performance on selected medical question-answering tasks \cite{singhal2023large,singhal2025toward}. More recent systems have demonstrated increasingly sophisticated conversational and differential-diagnostic behavior \cite{tu2025towards,mcduff2025towards},. These advances have encouraged proposals to use LLMs for patient communication, documentation, clinical decision support, and medical education. Nevertheless, strong performance on licensing-style examinations does not establish readiness for specialty practice. Examination questions are usually static, contain carefully selected information, and reward identification of one best answer. Real care instead requires the model to decide which information is reliable, what remains unknown, whether new evidence changes an earlier plan, and which errors could cause immediate harm.

Evidence from broader clinical evaluations has reinforced this concern. A systematic review of 519 health-care LLM studies found that accuracy dominated evaluation designs, whereas real patient-care data, calibration, fairness, workflow effects, and deployment considerations were uncommon \cite{bedi2025testing}. In a large evaluation based on intensive-care cases, LLMs showed limitations in instruction following, laboratory interpretation, guideline adherence, and robustness to the order and amount of clinical information \cite{hager2024evaluation}. A randomized clinical trial further demonstrated that simply giving clinicians access to an LLM did not necessarily improve diagnostic reasoning \cite{goh2024large}. Recent investigations and evaluation frameworks have distinguished correct final answers from the quality, consistency, and clinical usability of the reasoning process \cite{johri2025evaluation,rao2026large,tordjman2026limitations,qiu2025quantifying,agrawal2025evaluation}. These findings have motivated calls for benchmarks with stronger construct validity, scenario realism, clinically meaningful error taxonomies, and explicit safety assessment \cite{agrawal2025evaluation,zhou2025automating,croxford2025evaluating,zhang2025llmeval,wang2025novel,lekadir2025future,sounderajah2025stard}.

Respiratory medicine is particularly suited to scenario-based evaluation because it simultaneously demands longitudinal disease management, as exemplified by Chronic obstructive pulmonary disease (COPD), and multimodal diagnostic reasoning, as exemplified by pulmonary nodule assessment. COPD exemplifies the need for longitudinal assessment. Exacerbation risk depends on symptoms, prior exacerbations, medication use, oxygenation, lung function, comorbidities, and recent treatment response. Effective management therefore requires recognition of changes across visits and coordinated planning for inhaler optimization, adherence, rehabilitation, vaccination, monitoring, and escalation. The 2026 Global Initiative for Chronic Obstructive Lung Disease report provides current recommendations for diagnosis, risk assessment, prevention, and longitudinal treatment, while recent analyses have examined the growing role of artificial intelligence in integrated respiratory care \cite{GOLD2026,al2026ai}. Recent respiratory LLM studies have evaluated specialty outpatient diagnosis, spirogram interpretation, pulmonary nodule follow-up, and longitudinal CT assessment \cite{porcella2026large,bektacs2026evaluation,mei2026spirollm,wen2025evaluation,mao2025assessments}. However, a model can produce a plausible COPD summary while omitting a critical medication-safety check, failing to individualize nonpharmacological management, or recommending treatment changes without sufficient evidence.

Pulmonary nodule management presents a complementary multimodal challenge. Risk assessment depends on the CT appearance of the lesion, size, attenuation, morphology, multiplicity, interval change, clinical context, and the applicability of a specific management framework. Incidental nodules, screening-detected nodules, subsolid nodules, patients with previous cancer, and immunocompromised patients do not necessarily follow the same pathway \cite{woodhouse2025leveraging,macmahon2017guidelines,mazzone2018screening,society2024chinese,chen20242023,christensen2024acr}. The task is not merely to classify a nodule as benign or malignant. Clinicians must determine whether the available information supports surveillance, additional imaging, multidisciplinary review, tissue diagnosis, or surgery, while avoiding both delayed cancer evaluation and unnecessary invasive procedures. Recent studies have shown encouraging LLM performance in pulmonary nodule follow-up and longitudinal CT interpretation, but they have also identified incorrect intervals, unsupported feature descriptions, and potentially harmful recommendations \cite{wen2025evaluation,mao2025assessments}.

The emergence of multimodal LLMs increases both the potential value and the risk of respiratory decision support. Medical-image interpretation requires the model to ground language in visible regions and to distinguish observed findings from clinical inference. Reviews published in 2025 emphasize that multimodal systems remain vulnerable to hallucinated anatomy, weak localization, inconsistent reasoning, and poor calibration \cite{nam2025multimodal,rao2025multimodal}. In pulmonary nodule assessment, a response can appear clinically sophisticated while inventing spiculation, pleural retraction, vascular convergence, or interval growth that is not supported by the provided image. Such errors are especially problematic because subsequent recommendations may be logically coherent but founded on a fabricated premise. A benchmark therefore needs to capture not only whether the final management resembles a guideline, but whether the chain from image observation to risk interpretation and action is evidence-grounded.

Open-ended specialty evaluation also creates methodological challenges. Exact-match accuracy is unsuitable when several clinically acceptable formulations exist, yet an exclusively holistic score can reward fluency while overlooking omitted actions. Guideline-derived checklists and atomic clinical points improve transparency by identifying which diagnosis, monitoring, treatment, safety-netting, or escalation elements are explicitly present. Conversely, rubric-based evaluation can capture appropriateness, prioritization, internal consistency, and usability that are difficult to reduce to keyword matching. Recent work on physician-validated benchmarks and automated evaluators supports combining structured criteria with expert-calibrated judgment, while also warning that evaluator models require validation and should not be treated as infallible \cite{zhou2025automating,croxford2025evaluating,zhang2025llmeval,wang2025novel}. Safety must remain a separate analytical layer because the clinical consequences of an unsupported reassurance, dangerous medication adjustment, or invented imaging finding are not equivalent to a minor omission.

To address these gaps, we developed RESPClinBench, a real-world, scenario-based benchmark for respiratory clinical decision-making and whole-course disease management. Derived from de-identified clinical data, it preserves the complexity of real practice and evaluates complete care pathways from risk recognition and diagnosis to intervention, monitoring, and long-term follow-up. AECOPD-PIM assesses longitudinal COPD prediction and management, whereas PNBIM integrates chest CT images with structured clinical information for pulmonary nodule assessment. By combining complementary modalities and disease trajectories, RESPClinBench provides a clinically grounded framework for identifying model capabilities, task-specific limitations, and safety risks before integration into respiratory workflows.

\section{Method}
\subsection{Study design and benchmark scope}

RESPClinBench was designed as a scenario-based evaluation of clinical decision-making in respiratory specialty care. Rather than testing isolated factual recall, the benchmark represents the sequence of decisions required in actual practice, including recognition of risk, integration of multimodal or longitudinal evidence, selection of an appropriate diagnostic or therapeutic pathway, monitoring of treatment response, prevention of deterioration, and planning of continued care. Its three organizing domains are multimodal data-fusion diagnosis, chronic-disease risk prediction and intervention, and diagnostic compliance with medical-risk control. The present analysis used two open-ended datasets representing complementary respiratory workflows: longitudinal COPD exacerbation management and multimodal pulmonary nodule assessment (table \ref{tab:overview_of_dataset}).

\subsection{Real-world data adaptation and expert review}

All benchmark cases were adapted from de-identified real-world respiratory clinical data. The three attending-level respiratory physicians served as the primary clinical adapters rather than only as post hoc reviewers. For each source case, they reconstructed the clinically relevant chronology, selected the information required for the task, standardized the question, drafted and iteratively revised the expert reference answer, and decomposed the final answer into concise atomic clinical-action points. During revision, the physicians removed unsupported, redundant, overly broad, or ambiguous statements; reconciled recommendations with the supplied evidence and applicable guidelines; and ensured that every atomic point was independently answerable, clinically meaningful, and suitable for automated matching. They also checked that comorbidities, treatment exposure, uncertainty, and longitudinal changes from the original record were preserved without disclosing hidden labels. One senior respiratory specialist independently cross-reviewed the complete question-answer-action package and performed final adjudication. This review covered source fidelity, internal consistency, guideline concordance, clinical plausibility, answerability, safety, and completeness. Disagreements regarding case framing, reference-answer content, or atomic-point inclusion were resolved through revision, and cases with unresolved ambiguity or insufficient evidence were excluded. This multistage workflow was used to establish accurate and stable reference standards for both clinical content and automated evaluation.

\begin{table}
\centering
\caption{Overview of RESPClinBench Datasets and Evaluation Targets}
\label{tab:overview_of_dataset}
\begin{tabular}{>{\raggedright\arraybackslash}p{0.1\linewidth}>{\raggedright\arraybackslash}p{0.2\linewidth}>{\raggedright\arraybackslash}p{0.1\linewidth}>{\raggedright\arraybackslash}p{0.15\linewidth}>{\raggedright\arraybackslash}p{0.3\linewidth}}
\toprule
\textbf{Dataset} & \textbf{Clinical focus} & \textbf{Items} & \textbf{Input modality} & \textbf{Primary capabilities} \\
\midrule
AECOPD-PIM & AECOPD prediction, intervention, and whole-course management & 427 & Longitudinal text and structured clinical data & Risk warning; trend interpretation; medication review; monitoring; nonpharmacological intervention; escalation \\

PNBIM & Pulmonary nodule benign-malignant identification and dynamic management & 196 & Chest CT image plus demographic, exposure, laboratory, and history data & Image-grounded risk assessment; benign-malignant differentiation; diagnostic planning; treatment and follow-up management \\
\bottomrule

\end{tabular}

\end{table}

\subsection{AECOPD-PIM}

AECOPD-PIM contains 427 open-ended cases adapted from representative real-world respiratory outpatient trajectories. Cases span different degrees of baseline impairment and exacerbation risk and reproduce the clinical pathway from warning-signal detection to early intervention, treatment review, and ongoing disease management. Inputs may include repeated visits, symptom control, activity limitation, rescue medication, maintenance inhalers, antibiotics or systemic corticosteroids, oxygen saturation, pulmonary function, relevant laboratory findings, comorbidity, vaccination, rehabilitation, and antineoplastic or cardiovascular treatment. The expected response includes an individualized risk judgment, patient-specific evidence, early monitoring and intervention, medication review, nonpharmacological measures, and an escalation plan. The task therefore evaluates whether a model can maintain clinical continuity across visits rather than generate a one-time COPD summary. Reference actions were aligned with contemporary COPD guidance and specialist review \cite{GOLD2026,porcella2026large,wedzicha2017management}.

\subsection{PNBIM}

PNBIM contains 196 multimodal cases adapted from real pulmonary nodule data and covering lesions with different sizes, densities, imaging patterns, and clinical risk contexts. Each case combines a chest CT image with structured demographic, lifestyle, occupational exposure, laboratory, pulmonary-function, and medical-history information. The task requires image-grounded assessment of lesion nature, integration of clinical risk, selection of a reasonable diagnostic pathway, and planning of surveillance or intervention over time. Case construction avoided disclosure of pathology or other hidden gold-standard labels in the question. When a source image did not provide a specific morphological feature, that feature was not presented as an observed fact in the reference record. Recommendations were evaluated against applicable incidental-nodule, screening, subsolid-nodule, and expert-consensus frameworks \cite{macmahon2017guidelines,mazzone2018screening,society2024chinese,chen20242023,christensen2024acr}.

\subsection{Case format and reference standards}

Each record contained a clinical question and an expert reference answer. PNBIM additionally contained an image field. Reference answers were decomposed into concise clinical actions covering risk assessment, additional evaluation, treatment or surveillance, longitudinal monitoring, safety communication, and health management. These actions were selected for direct clinical relevance and reviewed to avoid redundancy. The three attending respiratory physicians performed primary adaptation and clinical verification, and the senior respiratory specialist cross-checked the final question-answer pairs and adjudicated disagreements. Evaluation criteria were tailored to each dataset while preserving common principles of factual grounding, guideline consistency, personalization, temporal coherence, whole-course management, and safety.

\subsection{Models and API inference protocol}

Seven models were evaluated: Gemini-3.5-Flash, Gemini-3.1-Pro-Preview, GPT-5.4, GPT-5.4-Mini, Grok-4.3, Qwen3.5-397B-A17B, and Qwen3.6-27B. All models were queried through application programming interfaces. Temperature was set to 0 and the maximum generation length was 8192 tokens for every model. Each benchmark item was submitted in an independent request to prevent conversational carryover. Models received the same dataset-specific task instruction and patient information, with image input included for PNBIM. The reference answer, clinical-action list, and evaluator annotations were not provided to the tested model. API failures were retried under the same inference settings; persistent failures were retained as unsuccessful outputs rather than silently removed.

\subsection{Evaluation framework}

RESPClinBench employed a dual-component automated evaluation framework that balanced explicit clinical-action coverage with holistic clinical-quality assessment. The first component measured coverage of guideline-derived atomic clinical-action points. For each response, every prespecified answer point was assessed independently and assigned a binary value of 1 when explicitly covered and 0 when absent. The item-level atomic-action recall was calculated as the number of recalled points divided by the total number of reference points for that item and was converted to a 0-to-100 scale. Dataset-level atomic-action performance was calculated as the unweighted mean of item-level recall scores. The second component used a structured LLM-as-a-Judge agent to assess responses against dataset-specific clinical rubrics covering medical accuracy, completeness, longitudinal consistency, guideline concordance, safety, communication, and direct clinical usability (table \ref{tab:rubric_dimension}). The judge score was likewise normalized to a 0-to-100 scale. The two components contributed equally to the composite score, with atomic-action recall accounting for 50\% and the LLM-as-a-Judge score accounting for 50\%; the final score was therefore their arithmetic mean. The automated judge also generated dimension-level scores and scenario-specific safety flags in a standardized machine-readable format.

\begin{table}
\centering
\caption{Dataset-specific rubric dimensions used for holistic clinical evaluation}
\label{tab:rubric_dimension}
\begin{tabular}{>{\raggedright\arraybackslash}p{0.2\linewidth}>{\raggedright\arraybackslash}p{0.7\linewidth}}
\toprule
\textbf{Dataset} & \textbf{Rubric dimensions} \\
\midrule
PNBIM & Lesion nature and risk assessment; clinical-imaging integration; diagnostic and staging planning; treatment-plan appropriateness; health management and safety communication; clarity and clinical usability \\

AECOPD-PIM & COPD warning and severity assessment; longitudinal course and medication-record understanding; follow-up and comprehensive assessment; medication adjustment and safety; nonpharmacological intervention; clarity and clinical usability \\
\bottomrule

\end{tabular}

\end{table}

\subsection{Safety and reliability outcomes}

PNBIM outputs were flagged for imaging hallucination when the response asserted an image feature or interval change that was not supported by the supplied input. Overdiagnosis indicated excessive diagnostic certainty or disproportionate escalation beyond the available evidence. A serious medical-risk flag identified recommendations with plausible potential for major harm. AECOPD-PIM outputs were evaluated for medication-safety risk, overdiagnosis, and serious medical risk. Risk rates were summarized by model, and final scores were compared between flagged and unflagged responses.

\subsection{Statistical analysis}

Mean, standard deviation, median, and bootstrap 95\% confidence intervals were calculated for final scores. Confidence intervals used 5,000 bootstrap resamples at the case level. Friedman tests compared the seven models on paired benchmark items. When the omnibus test was significant, paired Wilcoxon signed-rank tests were used for model comparisons, with Holm correction for multiplicity. Spearman correlation assessed stability of model rankings across PNBIM and AECOPD-PIM and the association between the two evaluation components. Mann-Whitney tests compared final scores between responses with and without each risk flag. Statistical tests were two-sided, and P<0.05 was considered significant.
\section{Result}
\subsection{Overall performance and cross-dataset stability}

\begin{figure}
    \centering
    \includegraphics[width=1\linewidth]{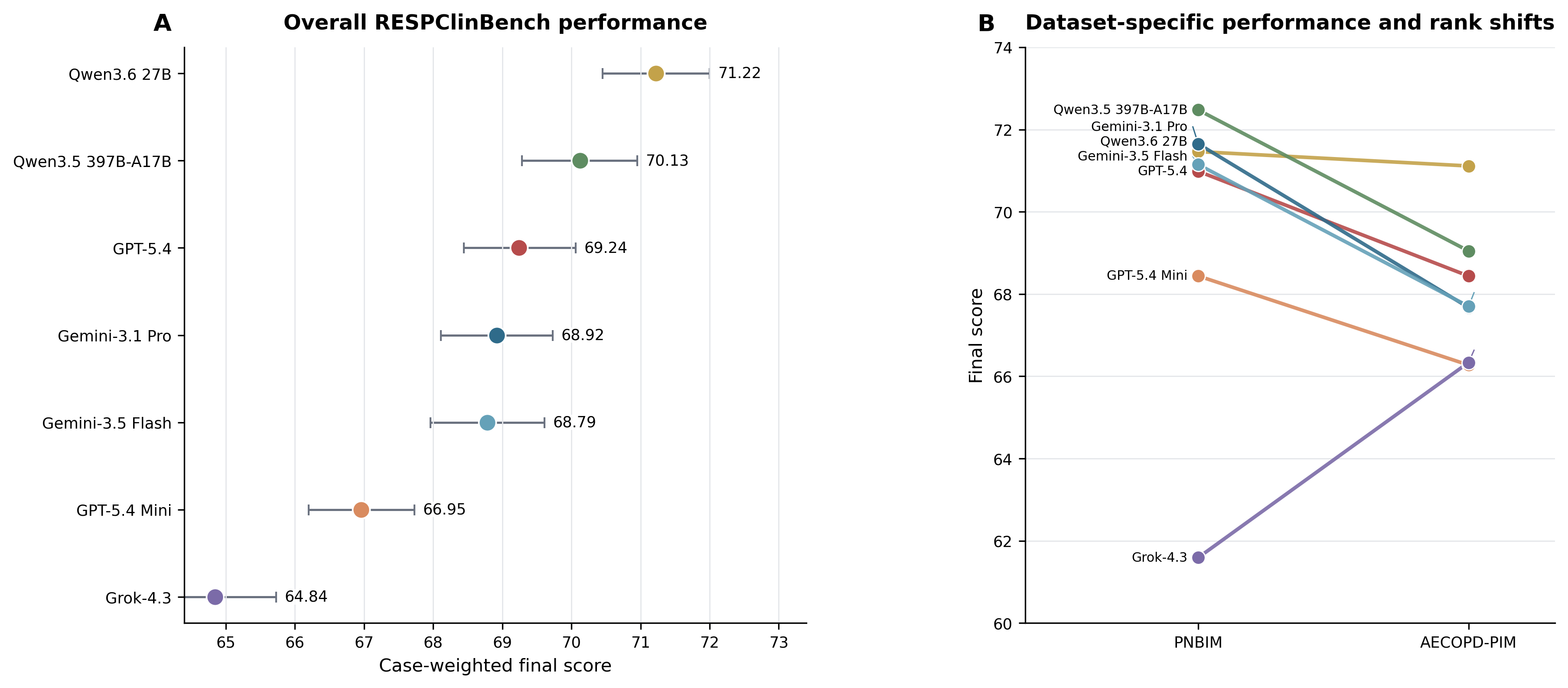}
    \caption{Overall and dataset-specific model performance in RESPClinBench. (A) Case-weighted final scores across all 623 cases; points indicate model means and horizontal error bars indicate 95\% bootstrap confidence intervals. (B) Final scores in PNBIM and AECOPD-PIM; each colored line represents one model and illustrates changes in relative performance between multimodal pulmonary nodule management and longitudinal COPD management. Higher scores indicate better performance. }
    \label{fig:fig1}
\end{figure}

The final analysis included 623 benchmark cases and seven models, yielding 4,361 responses: 1,372 for PNBIM and 2,989 for AECOPD-PIM. The mean final score across all model responses was 68.58. Qwen3.6-27B ranked first with a case-weighted score of 71.22 (95\% CI 70.47-71.98), followed by Qwen3.5-397B-A17B at 70.13 (95\% CI 69.28-70.94) and GPT-5.4 at 69.24 (95\% CI 68.45-70.07). Model differences were significant across the 623 paired cases (Friedman chi-square=380.63, P<0.001). The leading model exceeded the second-ranked model by 1.10 points (Holm-adjusted P=0.0016) and significantly outperformed every other model after correction. Overall model estimates and bootstrap confidence intervals are shown in Figure \ref{fig:fig1}A.

PNBIM had a mean final score of 69.68. Qwen3.5-397B-A17B ranked first at 72.48, followed by Gemini-3.1-Pro-Preview at 71.65 and Qwen3.6-27B at 71.46. AECOPD-PIM had a mean final score of 68.08; Qwen3.6-27B ranked first at 71.11, followed by Qwen3.5-397B-A17B at 69.05 and GPT-5.4 at 68.44. Differences between models were significant in both PNBIM (Friedman chi-square=202.49, P<0.001) and AECOPD-PIM (chi-square=256.41, P<0.001). Model ranks showed moderate cross-dataset correspondence, but the association was not statistically significant (Spearman rho=0.643, P=0.119), indicating that performance in multimodal nodule management did not fully predict performance in longitudinal COPD care. Dataset-specific rank shifts are visualized in Figure \ref{fig:fig1}B.

\subsection{Evaluation-component divergence and dimension-level performance}

\begin{figure}
    \centering
    \includegraphics[width=1\linewidth]{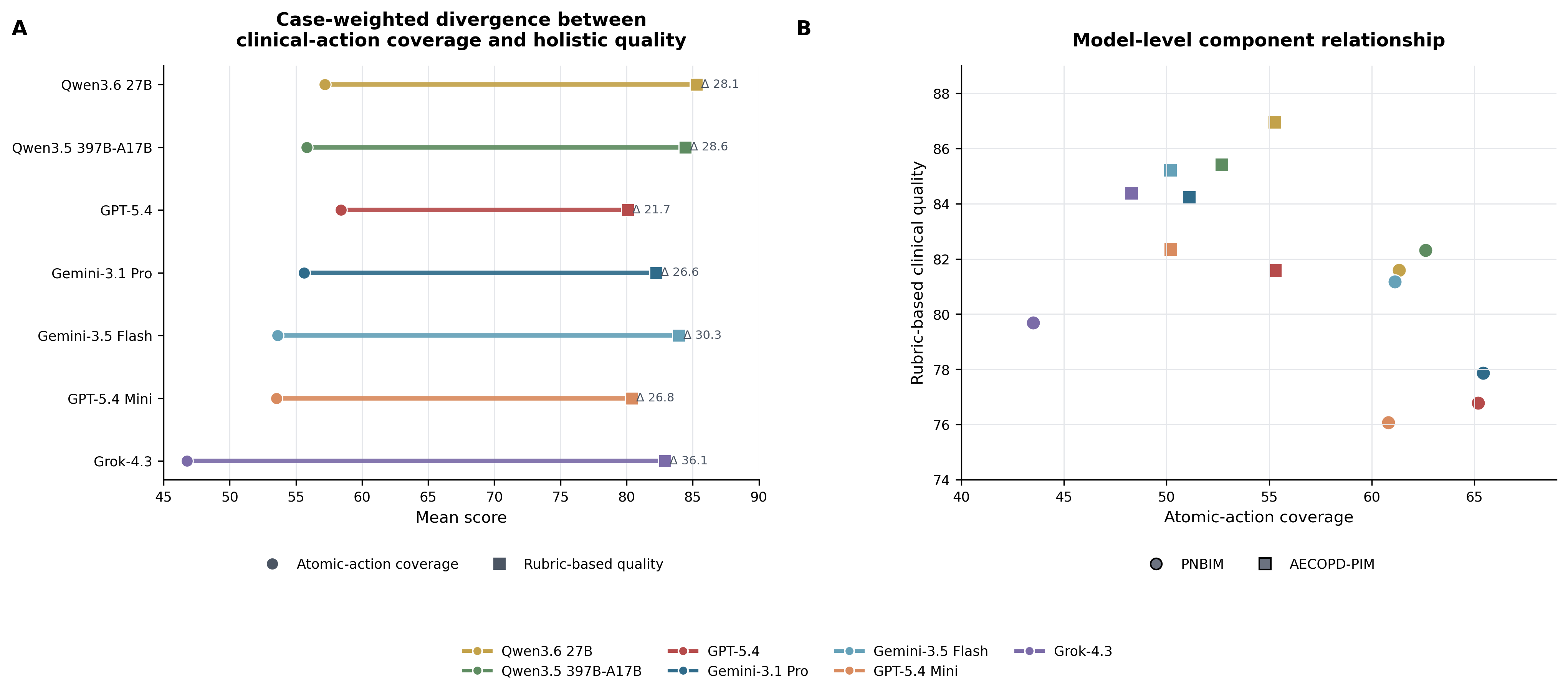}
    \caption{Divergence between guideline-derived clinical-action coverage and rubric-based clinical quality. (A) Case-weighted mean scores for the two evaluation components in each model; the labeled gap is the rubric-based score minus atomic-action coverage. (B) Model-level relationship between the two components in PNBIM and AECOPD-PIM. Circles denote PNBIM and squares denote AECOPD-PIM; colors identify models. The diagonal reference line represents equality between the components. Higher values indicate better performance.}
    \label{fig:fig2}
\end{figure}

The rubric-based clinical quality score was substantially higher than guideline-derived atomic-action coverage in both datasets. The mean gap was 19.36 points in PNBIM and 32.45 points in AECOPD-PIM. A gap exceeding 30 points occurred in 27.92\% of PNBIM responses and 55.07\% of AECOPD-PIM responses. Correlation between the two evaluation components was weak in PNBIM (Spearman rho=0.221, P<0.001) and weaker in AECOPD-PIM (rho=0.092, P<0.001). By contrast, atomic-action coverage was strongly correlated with the composite final score in both PNBIM (rho=0.926) and AECOPD-PIM (rho=0.912). These results indicate that a response could be judged coherent and clinically polished while omitting multiple explicit management actions. The case-weighted gaps and model-level component relationship are shown in Figure \ref{fig:fig2}.

\begin{figure}
    \centering
    \includegraphics[width=1\linewidth]{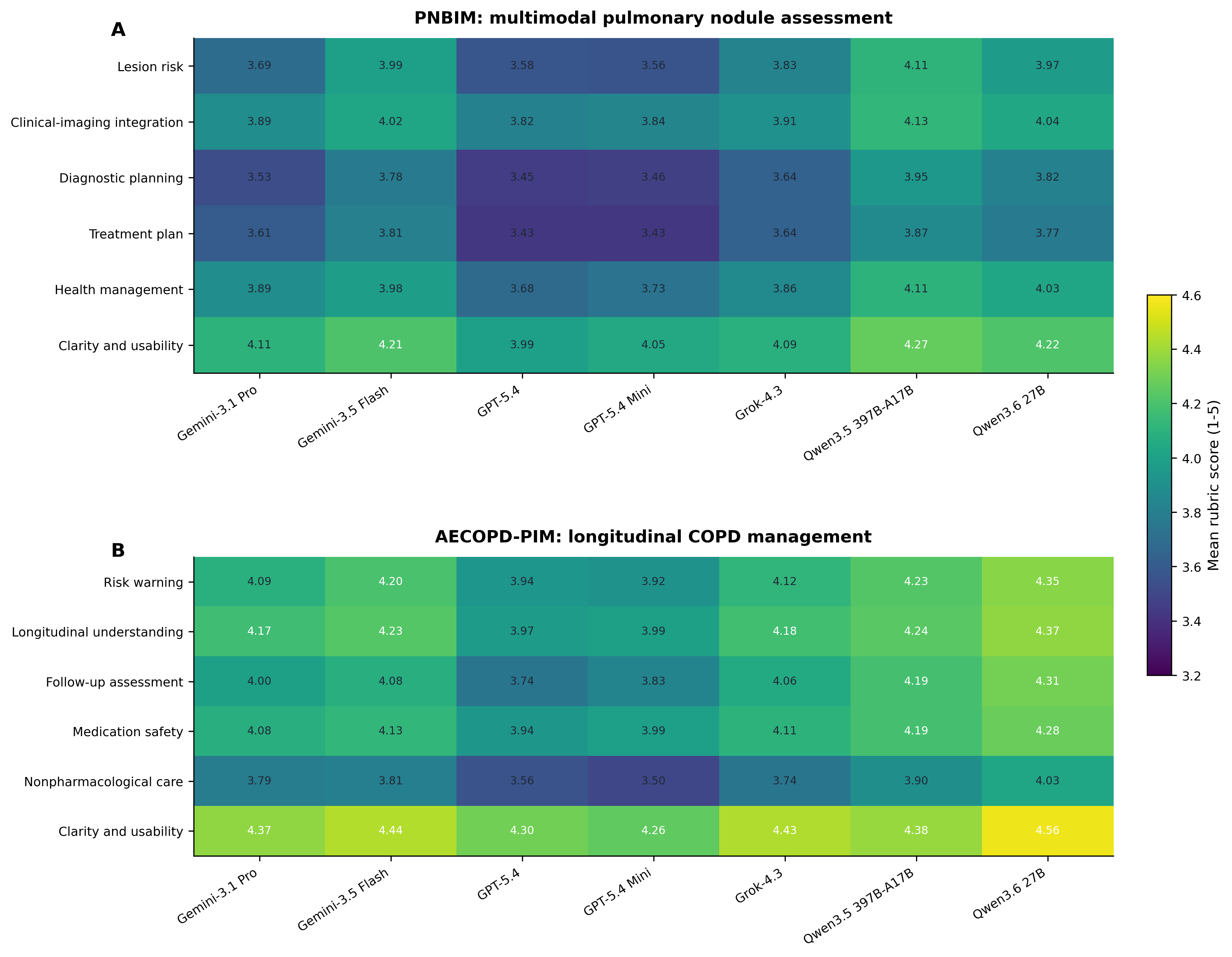}
    \caption{Dimension-level clinical performance across RESPClinBench models. (A) PNBIM rubric dimensions for multimodal pulmonary nodule risk assessment and management. (B) AECOPD-PIM rubric dimensions for longitudinal COPD warning, intervention, and management. Cell values are mean rubric scores on a 1-to-5 scale; darker shading indicates stronger performance. Dimensions are displayed using concise English labels, with full definitions provided in Table \ref{tab:rubric_dimension}.}
    \label{fig:fig3}
\end{figure}

Across PNBIM, clarity and clinical usability was the strongest dimension (mean 4.17/5), whereas treatment-plan appropriateness was the weakest (3.64/5), followed by diagnostic and staging planning (3.69/5). In AECOPD-PIM, clarity and clinical usability was again highest (4.39/5), while nonpharmacological intervention and health management was lowest (3.75/5). This pattern shows that answer presentation was consistently stronger than the operational completeness of diagnostic or management plans. The complete model-by-dimension profiles are presented in Figure \ref{fig:fig3}.
\subsection{Safety and reliability analysis}

\begin{figure}
    \centering
    \includegraphics[width=1\linewidth]{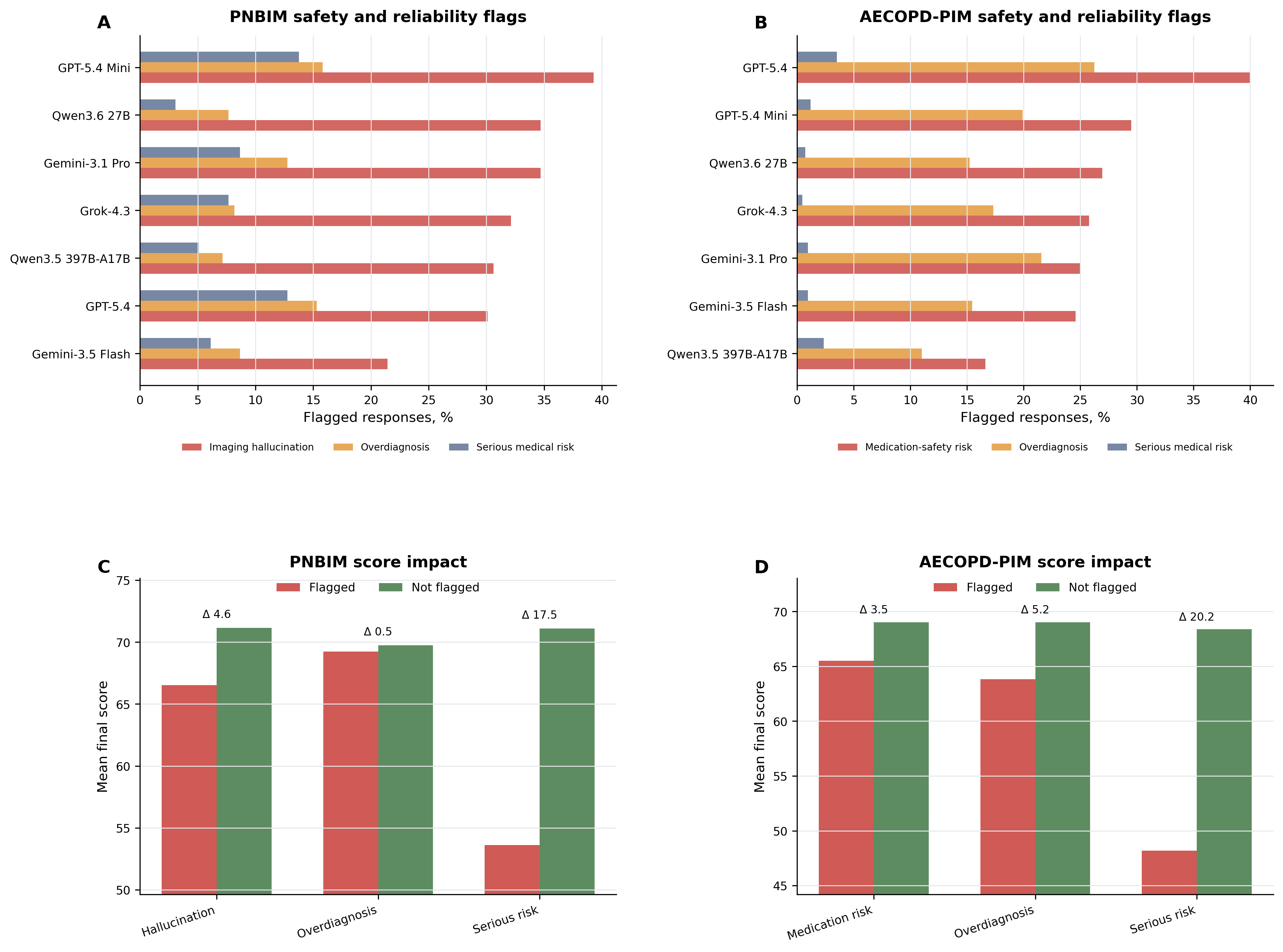}
    \caption{ Safety, reliability, and performance impact of scenario-specific risk flags. (A) Model-specific rates of imaging hallucination, overdiagnosis, and serious medical risk in PNBIM. (B) Model-specific rates of medication-safety risk, overdiagnosis, and serious medical risk in AECOPD-PIM. (C-D) Mean final scores for flagged and unflagged responses in PNBIM and AECOPD-PIM, respectively; delta values represent the unflagged-minus-flagged score difference. Risk categories were evaluated independently and were not mutually exclusive.}
    \label{fig:fig4}
\end{figure}

Among 1,372 PNBIM responses, imaging hallucination was identified in 437 (31.85\%), overdiagnosis in 148 (10.79\%), and serious medical risk in 112 (8.16\%). Imaging hallucination varied from 21.43\% for Gemini-3.5-Flash to 39.29\% for GPT-5.4-Mini. Serious-risk responses scored 17.48 points lower than unflagged responses (53.62 vs 71.11, P<0.001), and imaging-hallucination responses scored 4.62 points lower (66.53 vs 71.15, P<0.001). Final scores did not differ significantly between responses with and without the overdiagnosis flag (P=0.376). Model-specific rates and their association with final scores are shown in Figure \ref{fig:fig4}A and Figure \ref{fig:fig4}C.

Among 2,989 AECOPD-PIM responses, medication-safety risk was flagged in 805 (26.93\%), overdiagnosis in 541 (18.10\%), and serious medical risk in 43 (1.44\%). Medication-safety risk ranged from 16.63\% for Qwen3.5-397B-A17B to 40.05\% for GPT-5.4. Serious-risk responses scored 20.19 points lower than unflagged responses (48.18 vs 68.37, P<0.001). Medication-safety flags were associated with a 3.51-point reduction (65.52 vs 69.03, P<0.001), and overdiagnosis with a 5.19-point reduction (63.83 vs 69.02, P<0.001). Model-specific rates and score comparisons are shown in Figure \ref{fig:fig4} and Figure \ref{fig:fig4}D.

\section{Discussion}
\subsection{Principal findings}
RESPClinBench evaluated seven contemporary models across 623 respiratory cases representing two distinct but complementary workflows. Unlike static examination datasets, these cases were adapted from real respiratory clinical data and required models to complete clinically meaningful pathways, including risk recognition, evidence integration, intervention planning, monitoring, and longitudinal management. Overall performance was moderate, and the highest score was 71.22 rather than approaching ceiling performance. No model led both datasets: Qwen3.5-397B-A17B ranked first for pulmonary nodule management, while Qwen3.6-27B ranked first for COPD warning and longitudinal management. Cross-dataset rank stability was only moderate. These findings support the central premise that respiratory competence is task dependent and cannot be inferred from a general medical leaderboard or a single specialty subset.

A persistent gap remained between rubric-based clinical quality and atomic clinical-action coverage, particularly in AECOPD-PIM, where the mean difference exceeded 32 points. The holistic rubric rewarded plausibility, coherence, prioritization, and usability, whereas atomic-action recall required each prespecified element to be stated explicitly. Models could therefore produce fluent high-level plans while omitting specific actions related to monitoring, medication review, rehabilitation, prevention, or escalation. This discrepancy was amplified in longitudinal COPD care, which involves numerous sequential and patient-specific requirements. Because the two components contributed equally to the final score, fluent presentation could not compensate for incomplete action coverage. This pattern reflects the evaluation illusion described previously \cite{agrawal2025evaluation} and supports combining structured action-level assessment with rubric-based clinical judgment.

\subsection{Multimodal pulmonary nodule management}
PNBIM required the model to link CT information with demographic and clinical context and then translate that assessment into a management plan. Imaging hallucination occurred in 31.85\% of responses, showing that access to multimodal input did not consistently produce image-grounded reasoning. Existing studies have reported useful LLM performance in pulmonary nodule follow-up and longitudinal CT interpretation \cite{wen2025evaluation,mao2025assessments}, but the present findings show that unsupported descriptions of morphology or interval change remain common when image interpretation is embedded within a broader clinical response.

Serious medical risk occurred in 8.16\% of PNBIM outputs, or approximately one in 12 responses. Although less frequent than imaging hallucination, these cases involved recommendations with plausible potential for major harm and scored 17.48 points lower than unflagged outputs. Unsupported imaging descriptions may distort malignancy assessment and lead to inappropriate reassurance, delayed evaluation, unnecessary invasive testing, or premature surgery. Imaging hallucination and serious medical risk are therefore related but distinct: the former reflects poor grounding, whereas the latter reflects the potential clinical consequences of the recommendation \cite{nam2025multimodal,rao2025multimodal}.

Treatment-plan appropriateness and diagnostic or staging planning were the weakest PNBIM dimensions. These tasks require more than recognizing malignancy-associated morphology. The model must determine which guideline is applicable, whether comparison imaging is available, whether surveillance is sufficient, and whether additional testing or invasive intervention is proportionate. Incorrectly applying an incidental-nodule pathway to a screening case, or treating a high-risk imaging impression as a pathological diagnosis, can generate either delayed care or unnecessary intervention. The finding that overdiagnosis did not consistently reduce the composite score further supports retaining explicit safety flags rather than assuming that a single total score captures all clinically important behavior.

\subsection{Longitudinal COPD care}
AECOPD-PIM exposed a different weakness: models generally produced readable assessments but incompletely operationalized long-term care. Nonpharmacological intervention was the lowest-scoring dimension, despite pulmonary rehabilitation, vaccination, physical activity, exposure reduction, nutritional assessment, self-management, and individualized follow-up being central to COPD outcomes \cite{GOLD2026,porcella2026large,wedzicha2017management}. A model that recommends inhaler escalation but omits adherence, inhaler technique, rehabilitation, infection prevention, and escalation thresholds does not reproduce the whole-course decision process expected in respiratory practice.
Medication-safety flags affected more than one quarter of AECOPD-PIM outputs. These concerns included treatment changes without adequate context, insufficient attention to concurrent medication and comorbidity, and failure to distinguish maintenance optimization from emergency management. The association between medication-safety flags and lower final scores confirms that the evaluation framework captured part of this risk, but the high absolute frequency remains concerning. Recent respiratory LLM work has emphasized that clinical integration requires constrained use cases, human oversight, and transparent limits \cite{porcella2026large,bektacs2026evaluation,mei2026spirollm,wen2025evaluation,mao2025assessments}. RESPClinBench complements these studies by providing a reproducible way to quantify treatment-related risk in longitudinal scenarios.
\subsection{Implications for benchmark design and clinical deployment}
The results support a multidimensional approach to specialty evaluation. Structured clinical-action coverage identifies omissions and facilitates error analysis. Rubric-based assessment captures prioritization, coherence, guideline appropriateness, and usability. Safety flags preserve information that may be diluted in an aggregate score. Dimension-level analyses show whether a model succeeds because it writes clearly or because it actually performs the required clinical work. Similar principles underpin recent physician-validated benchmarks, clinical evaluator studies, and safety-effectiveness frameworks \cite{zhou2025automating,croxford2025evaluating,zhang2025llmeval,wang2025novel,lekadir2025future,sounderajah2025stard}.
For deployment, model selection should be matched to the intended respiratory task. A system used to draft follow-up instructions requires different validation from one that interprets CT images or recommends inhaler changes. Institutions should evaluate the exact model version and API configuration, maintain human review, log inputs and outputs, monitor critical errors, and repeat validation after model updates. Prospective testing should begin in silent or simulated workflows and follow contemporary reporting and governance guidance \cite{lekadir2025future,sounderajah2025stard,wedzicha2017management,vasey2022reporting,collins2024tripod+,hernandez2020minimar}. The scores reported here do not justify autonomous diagnosis or treatment; they characterize model behavior under controlled benchmark conditions.
\subsection{Strengths and limitations}
The study has several strengths. RESPClinBench is explicitly oriented toward real respiratory clinical scenarios rather than generic examination questions. Its cases were adapted from de-identified real-world data, allowing the benchmark to retain clinically relevant comorbidity, treatment history, temporal change, and uncertainty. It evaluates both multimodal pulmonary nodule management and longitudinal COPD care, thereby testing not only diagnosis but also disease-course management from risk recognition through intervention and follow-up. Dataset construction was performed by three attending-level respiratory physicians and cross-reviewed and adjudicated by one senior respiratory specialist, providing a structured safeguard for source fidelity, clinical accuracy, guideline concordance, and reference-answer stability. The benchmark also uses expert-derived clinical actions, separately measures holistic quality and explicit content, and reports scenario-specific safety outcomes. Removing a single unsupported model series and recalculating all results ensured that rankings and risk estimates reflected a consistent seven-model comparison. The analysis further moves beyond overall scores by examining component concordance, cross-dataset transfer, dimension profiles, and risk-associated score differences.
Several limitations should be considered. The benchmark is retrospective and does not measure patient outcomes or clinician behavior. PNBIM uses selected CT images rather than the complete interactive image stack available to a radiologist, which may affect both model and human interpretation. Some histories are standardized to create controlled clinical contexts and may be more internally coherent than routine records. The evaluation agent, although structured, can still introduce systematic bias; human validation remains important. Risk flags capture prespecified error types and do not exhaust all possible harms. Model providers may update systems without changing public names, so results are time- and configuration-dependent. Finally, the benchmark currently contains two datasets and should be expanded to asthma, pulmonary function, infection, interstitial lung disease, bronchiectasis, respiratory failure, and other representative workflows.
\section{Conclusion}
RESPClinBench demonstrates the value of evaluating large language models within realistic respiratory workflows rather than relying on static knowledge questions. Built from real-world respiratory data and curated through multistage review by attending and senior respiratory specialists, the benchmark tests whether models can integrate evidence, maintain clinical continuity, and produce safe decisions across the course of disease management. Multimodal pulmonary nodule management was limited by frequent unsupported imaging interpretation and imperfect treatment planning, while longitudinal COPD care was limited by incomplete nonpharmacological management and medication-safety concerns. Model rankings changed across the two datasets, and no model achieved uniformly strong performance. Specialty evaluation should therefore combine explicit clinical-action coverage, expert-style assessment, dimension-level analysis, and independent safety outcomes. RESPClinBench provides a clinically grounded and reproducible framework for measuring these capabilities before LLMs are introduced into respiratory decision-support workflows.
\section{Data Availability}
The RESPClinBench d dataset is publicly available on the MedBench(\href{https://medbench.opencompass.org.cn/track-detail/11}{https://medbench.opencompass.org.cn/track-detail/11}) platform for model evaluation.

\bibliographystyle{unsrt}  
\bibliography{references}

\end{document}